\documentclass[10pt,twocolumn,letterpaper]{article}

\usepackage{cvpr}
\usepackage{times}
\usepackage{epsfig}
\usepackage{graphicx}
\usepackage{amsmath}
\usepackage{amssymb}
\usepackage{subcaption}
\usepackage{multirow}
\graphicspath{ {figures/} }
\renewcommand{\figurename}{Figure }

\usepackage[pagebackref=true,breaklinks=true,letterpaper=true,colorlinks,bookmarks=false]{hyperref}

 \cvprfinalcopy 
\usepackage{balance}

\def\cvprPaperID{****} 

\ifcvprfinal\fi
\begin{document}

\title{VidLoc: A Deep Spatio-Temporal Model for 6-DoF Video-Clip Relocalization}

\author{Ronald Clark\textsuperscript{1},
Sen Wang\textsuperscript{1},
Andrew Markham\textsuperscript{1},
Niki Trigoni\textsuperscript{1}, Hongkai Wen\textsuperscript{2}\\
\textsuperscript{1}University of Oxford, Oxford, OX1 3PA\\
\textsuperscript{2}University of Warwick, Coventry, CV4 7AL\\
{\tt\small firstname.lastname@cs.ox.ac.uk}
}

\maketitle

\begin{abstract}
Machine learning techniques, namely convolutional neural networks (CNN) and regression forests, have recently shown great promise in performing 6-DoF localization of monocular images. However, in most cases image-sequences, rather only single images, are readily available. To this extent, none of the proposed learning-based approaches exploit the valuable constraint of temporal smoothness, often leading to situations where the per-frame error is larger than the camera motion. In this paper we propose a recurrent model for performing 6-DoF localization of video-clips. We find that, even by considering only short sequences (20 frames), the pose estimates are smoothed and the localization error can be drastically reduced. Finally, we consider means of obtaining probabilistic pose estimates from our model. We evaluate our method on openly-available real-world autonomous driving and indoor localization datasets. 
\end{abstract}

\section{Introduction}

Localization of monocular images is a fundamental problem in computer vision and robotics. Camera localization forms the basis of many functions in computer vision where it is an important component of the Simultaneous Localization and Mapping (SLAM) process and has direct application, for example, in the navigation of autonomous robots and drones in first-response scenarios or the localization of wearable devices in assistive living applications. 

The most common means of performing 6-DOF pose estimation using visual data is to make use of specially-built models, which are constructed from a vast number of local features that have been extracted from the images captured during mapping. The 3D locations of these features are then found using a Structure-from-Motion (SfM) process, creating a many-to-one mapping from feature descriptors to 3D points. Traditionally, localizing a new query image against these models involves finding a large set of putative correspondences. The pose is then found using RANSAC to reject outlier correspondences and optimize the camera pose on inliers. Although this traditional approach has proven to be incredibly accurate in many situations, it faces numerous and significant challenges. These methods rely on local and unintuitive hand-crafted features, such as SIFT keypoints. Because of their local nature, establishing a sufficient number of reliable correspondences between the image pixels and the map is very challenging. Spurious correspondences arise due to both ``well-behaved'' phenomena such as sensor noise and quantization effects as well as pure outliers which arise due to the local correspondence assumptions not being satisfied \cite{goldstein2016shapefit}. These include inevitable environmental appearance changes due to, for example, changing light levels or dynamic elements such as clutter or people in the frame or the opening and closing of doors. These aspects conspire to give rise to a vast number of suprious correspondences, making it difficult to use for any purpose but the localization of crisp and high-resolution images. Secondly, the maps often consists of millions of elements which need to be searched, making it very computationally intensive and difficult to establish correspondences in real-time. 

\begin{figure}[h!]
	    \centering
	        \includegraphics[width=\columnwidth]{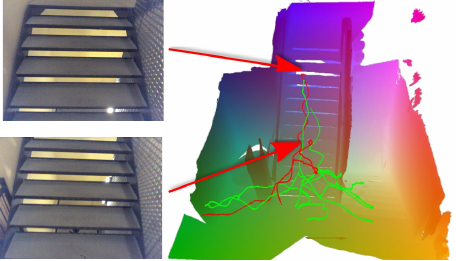}

	    \caption{An extreme example of perceptual aliasing in the Stairs scene of the Microsoft 7-Scenes dataset. One of the frames is taken at the bottom of the staircase and the other near the top. Using only single frames, as in the competing approaches, it would be impossible to correctly localize these images. }
	    \label{fig:Exp7ScenesImages}
	\end{figure}
	
Recently, however, it has been shown that machine learning methods such as random forests \cite{shotton2013scene} and convolutional neural networks (CNNs) \cite{kendall2015posenet} have the ability to act as a regression model which directly estimates pose from an input image with no expensive feature extraction or feature matching processes required. These methods consider the input images as being entirely uncorrelated and produce independent pose estimates that are incredibly noisy when applied to image sequences. On most platforms, including smart-phones, mobile robots and drones, image-sequences are readily obtained and have the potential to greatly enhance the accuracy of these approaches and promising results have been obtained for sequence-based learning for relative pose estimation \cite{clark17vinet}. Therefore, in this paper we consider ways in which we can leverage the temporal smoothness of image-sequences to improve the accuracy of 6-DoF camera re-localization. Furthermore, we show how we can in essence unify map-matching, model-based localization, and temporal filtering all in one, extremely compact model.

\subsection{Related Work}

\noindent \textbf{Map-matching} 
Map matching methods make use of a map of a space either in the form of roads and traversable paths or a floor-plan of navigable and non-navigable areas to localize a robot as it traverses the environment. Map-matching techniques are typified by their non-reliance on strict data-association and can use both exteroceptive (eg. laser scans) or interoceptive (odometry, the trajectory or the motion of the platform) sensors to obtain a global pose estimate. The global pose estimate is obtained through probabilistic methods such as sequential Monte Carlo (sMC) filtering \cite{gustafsson2002particle} or hidden Markov models (HMMs) \cite{newson2009hidden}. These methods inherently incorporate sequential observations, but accuracy is inferior to localizing against specialised maps, such as a 3D map of sparse features.

\noindent \textbf{Sparse feature based localization} 
When a 3D model of discriminative feature points is available (eg. obtained using SfM) then the poses of query images can be found using camera re-sectioning. Matching against large 3D models is generally very computationally expensive and requires lots of memory space to store the map. A number of approaches have been proposed to improve the efficiency of standard 3D-to-2D feature matching between the image and the 3D model \cite{zhang2006image}. For example, \cite{sattler2011fast} propose a quantized feature vocabulary for direct 2D-to-3D matching with the camera pose being found using RANSAC in combination with a PnP algorithm and in \cite{sattler2012improving} an active search method is proposed to efficiently find more reliable correspondences. \cite{middelberg2014scalable} propose a client-server architecture where the client exploits sequential images to perform high-rate local 6-DoF tracking which is then combined with lower-rate global localization updates from the server, entirely eliminating the need for loop-closure. The authors propose various methods to integrate the smooth local poses with the global updates. While this approach exploits a temporal stream of images, it does not use this information to improve the accuracy of global localization. In \cite{kroeger2014video} the authors consider means of improving the global accuracy by introducing temporal constraints into the image registration process by regularizing the poses trough smoothing.

\noindent \textbf{CNN features}
Deep learning is quickly becoming the dominant approach in computer vision. The many layers of a pre-trained CNN form a hierarchical model with increasingly higher level representations of the input data as one moves up the layers. It has been shown that many computer vision related tasks benefit from using the output from these upper layers as feature representations of the input images. These features have the advantage of being low-level enough to provide representations for a large number of concepts, yet are abstract enough to allow these concepts to be recognized using simple linear classifiers \cite{sharif2014cnn}. They have shown great success applied to a wide range of tasks including logo classification \cite{bianco2015logo}, and more closes related to our goals, scene recognition \cite{zhou2014learning} and place recognition \cite{sunderhauf2015place}.  

\noindent \textbf{Posenet} \cite{kendall2015posenet} demonstrated the feasibility of estimating the pose of a single RGB image by using a deep CNN to regress directly on the pose. For practical camera relocalization, Posenet is far from ideal. For example, on the Microsoft 7-Scenes dataset it achieves a $0.48m$ error where the model space is only $2.5m \times 1m \times 1m$. We argue that this can be partly attributed to the fact that the design of the network makes no attempt to capture the structure of the scene or map in which it is attempting to localize and thus would require an excessively large and well sampled training set to generalize adequately. 

\noindent \textbf{Scene coordinate regression forests} of Shotton et al. \cite{shotton2013scene} use a regression forest to learn the mapping between the pixels of an RGB-D input image and the scene co-ordinates of a previously established model. In essence the regression forest learns the function $f: (r,g,b,d,u,v) \rightarrow (U,V,W)$. To perform localization, a number of RGB-D pixels from the query image are fed through the forest and a RANSAC-based pose computation is used to determine a consistent and accurate final camera pose. To account for the temporal regularity of image sequences, the authors consider a frame-to-frame extension of their method. To accomplish this, they initialize one of the pose hypotheses with that obtained from the previous frame, which results in a significant improvement in localization accuracy. Although extremely accurate, the main disadvantage of this approach is that it requires depth images to function and does not eliminate the expensive RANSAC procedure. 

\subsection{Contributions}

In this paper, we propose a recurrent model for reducing the pose estimation error by using multiple frames for the pose prediction. Our specific contributions are as follows:

\begin{enumerate}
   \item We present a CNN-Recurrent Neural Network (RNN) model for efficent global localization from a monocular image sequence.
    \item We integrate into our network a method for obtaining the instantaneous covariances of pose estimates.
    \item We evaluate our approach to two large open datasets and answer the important question: How does our method compare to simply smoothing pose estimates as a post-processing step? 
\end{enumerate}

\section{Proposed Model}

\begin{figure}[t!]
    \centering
    \includegraphics[width=\columnwidth]{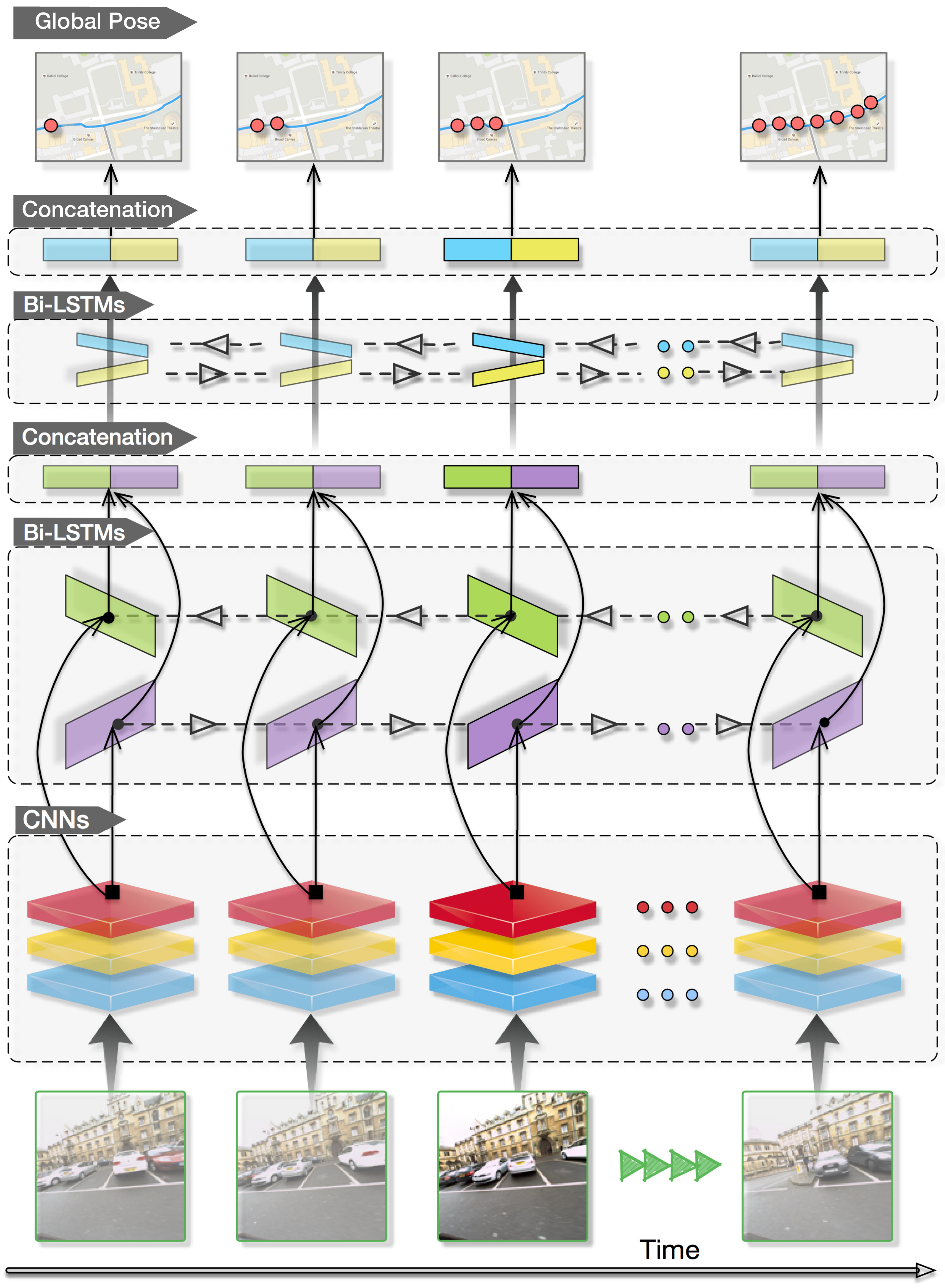}
    \caption{The CNN-RNN network for video-clip localization.}
    \label{fig:vidloc_archi}
\end{figure}

In this section we outline our proposed model for video-clip localization, VidLoc, a high-level overview of which is shown in Figure \ref{fig:vidloc_archi}. Our model processes the video image frames using CNN and integrates temporal information through a bidirectional LSTM.

\subsection{Image Features: CNN}
The goal of the CNN part of our model is to extract relevant features from the input images that can be used to predict the global pose of an image. A CNN consists of stacked layers performing convolution and pooling operations on the input image. There are a large number of CNN architectures that have been proposed, most for classifying objects in images and trained on the Imagenet database. These models, however, generalize well to other tasks, including pose estimation. As in the Posenet \cite{kendall2015posenet} paper, VGGNet \cite{simonyan2014very} is able to produce more accurate pose estimates, but incurs a high-computational cost due to its very deep architecture. As we are interested in processing multiple images in a temporal sequence we adopt the GoogleNet Inception \cite{szegedy2015going} architecture for the VidLoc CNN. We use only the convolutional and pooling layers of GoogleNet and drop all the fully-connected layers. In our experiments, we explore the impact on computational efficiency incurred vs. the increase in accuracy obtained using multiple frames.

\subsection{Temporal Modelling: Bidirectional RNN}
In Posenet and many other traditional image based localization approaches, the pose estimates are produced entirely independently for each frame. However, when using image-streams with temporal continuity, a great deal of pose information can be gained by exploiting the temporal regularity. For example, adjacent images often contain views of the same object which can boost the confidence in a particular location, and there are also tight constraints on the motion that can be undergone in-between frames - a set of frames estimated to be at a particular location are very unlikely to contain one or two located far away. 

\begin{figure}[h!]
    \centering
    \includegraphics[width=0.7\columnwidth]{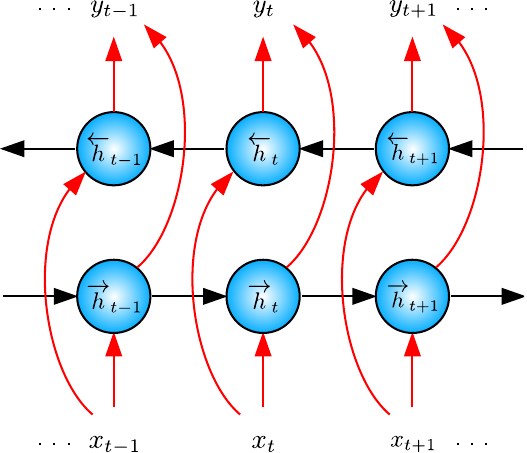}
    \caption{The structure of a bidirectional RNN \cite{schuster1997bidirectional}.}
    \label{fig:vidloc_model}
\end{figure}

To capture these dynamic dependencies, we make use of the LSTM model in our network. The LSTM \cite{hochreiter1997long} extends standard RNNs to enable them to learn long-term time dependencies. This is accomplished by including a \emph{forget gate}, input and output \emph{reset gates} and a \emph{memory cell}. The flow of information into and out-of the memory cell is regulated by the forget and input gates. This allows the network to overcome the vanishing gradient problem during training and thereby allow it to learn long-term dependencies. The input to the LSTM is the output of the CNN consisting of a sequence of feature vectors, $\mathbf{x}_{t}$. The LSTM maps the input sequence to the output sequence consisting of the global pose parameterised as a 7-dimensional vector, $\mathbf{y}_{t}$ consisting of a translation vector and orientation quaternion. The activations of the LSTM are computed by interatively applying the following operations on each timestep
\begin{equation}
\begin{aligned}
f_{t}&=\sigma _{g}(W_{f}\mathbf{x}_{t}+U_{f}\mathbf{h}_{t-1}+b_{f})\\
i_{t}&=\sigma _{g}(W_{i}\mathbf{x}_{t}+U_{i}\mathbf{h}_{t-1}+b_{i})\\
o_{t}&=\sigma _{g}(W_{o}\mathbf{x}_{t}+U_{o}\mathbf{h}_{t-1}+b_{o})\\
c_{t}&=f_{t}\circ c_{t-1}+i_{t}\circ \sigma _{c}(W_{c}\mathbf{x}_{t}+U_{c}\mathbf{h}_{t-1}+b_{c})\\
\mathbf{h}_{t}&=o_{t}\circ \sigma _{h}(c_{t}) \\
\mathbf{y}_t &=\sigma _{o}( W_{y} h_{t} + b_y)
\label{eqn:lstm}
\end{aligned}
\end{equation}
where $W, U$ and $b$ are the parameters of the LSTM, $f_t, i_t, o_t$ are the gate vectors, $\sigma _{g}$ is the non-linear activation function and $h_t$ is the hidden activation of the LSTM. For the inner activations, we use a hyperbolic tangent function and for the output $\sigma _{o}$ we use a linear activation. A limitation of the standard LSTM model is that it is only able to make use previous context in predicting the current output. For our monocular image-sequence pose prediction application we have a sliding window of frames available at any one instance in time and thus we can exploit both future and past contextual information predicting the poses for each frame in the sequence. For this reason, we adopt a Bidirectional architecture \cite{schuster1997bidirectional} for our LSTM model. The bidirectional model assumes the same state equations as in \ref{eqn:lstm}, but uses both future and past information for each frame by using two hidden states, $\overleftarrow{\mathbf{h}}_t$ and $\overrightarrow{\mathbf{h}}_t$, one for processing the data forwards and and the other for processing backwards, as shown in Figure \ref{fig:vidloc_model}. The hidden states are then combined to form a single hidden state $\mathbf{h}_t$ through a concatenation operation  
\begin{equation}
 \mathbf{h}_t = \left[\overleftarrow{\mathbf{h}_t},\overrightarrow{\mathbf{h}_t}\right]
\end{equation}
The output pose is computed from this hidden layer as in \ref{eqn:lstm}.

\subsection{Network Loss}
\label{sec:loss}

In order to train the network we use the sum of the Euclidean error magnitude of both the translation and orientation. To compute the loss, we separate the output of the LSTM into the translation $\mathbf{x}_t$ and orientation $\mathbf{q}_t$
\begin{equation}
\mathbf{y}_t = \left[\mathbf{x}_t,\mathbf{q}_t\right]
\end{equation}
and use a weighted sum of the error magnitudes of the two component vectors

\begin{equation}
    \mathcal{L}  = \sum_{t=1}^{T} \alpha_1 ||\mathbf{x}_t - \hat{\mathbf{x}}_t|| + \alpha_2 ||\mathbf{q}_t - \hat{\mathbf{q}}_t||
\end{equation}

We propagate the loss through the temporal frames in each training sequence by unrolling the network and performing back-propagation through time. To update the weights of the layers, we make use of the Adam optimizer.

\subsection{Probabilistic Pose Estimates}

Pose estimation methods, no matter how accurate, will always be subject to a degree of uncertainty. Being able to correctly model and predict uncertainty is thus a key component of any useful visual localization method. The euclidean sum-of-squares error which we defined in Sec. \ref{sec:loss} results in a network which approximates only the uni-modal conditional mean of the pose as defined by the training data. In essence the output of the network can be regarded as predicting $\mu_{\mathbf{x}}$, the mean of the conditional pose distribution $p\left([\mathbf{x,q}]|I\right) = \mathcal{N}\left(\mu_{[\mathbf{x,q}]},\sigma\right)$ where the Guassian assumption is induced by the use of the square error loss. For the unlikely case where the actual posterior pose distribution is Gaussian, this mean represents the optimal distribution in a maximum-likelihood sense. However, for global camera re-localization as we are concerned with in this paper, this assumption is unlikely. In many instances the appearance of a space is similar at multiple locations, for example, two corridors in a building may appear very similar (known as the ``perceptual aliasing'' problem and in most instances cannot be addressed using visual data alone). For this reason, we 

In \cite{kendall2015modelling}, one possible means of representing uncertainty in the global pose estimation was considered. In this work, the authors create a Bayesian convolutional neural network by using dropout as a means of sampling the model weights. The posterior distribution of the model weights $p\left( \mathbf{W}| \mathbf{X},\mathbf{Y} \right)$ is intractable and they use variational inference to approximate it as proposed in \cite{gal2015bayesian}. To produce probabilistic pose estimates, Monte Carlo pose samples are drawn and the mean and variance determined from these. Although this models the uncertainty in the \emph{model weights} correctly (i.e. the distribution of the model weights according to the training data), it does not fully capture the uncertainty of the pose estimates.

To model the pose uncertainty, we adopt the mixture density networks method \cite{bishop1994mixture}. This approach replaces the Gaussian with a mixture model, allowing a multi-modal posterior output distribution to be modelled. Using this approach, the pose estimates now take the form 

\begin{figure}[h!]
    \centering
    \includegraphics[width=\columnwidth]{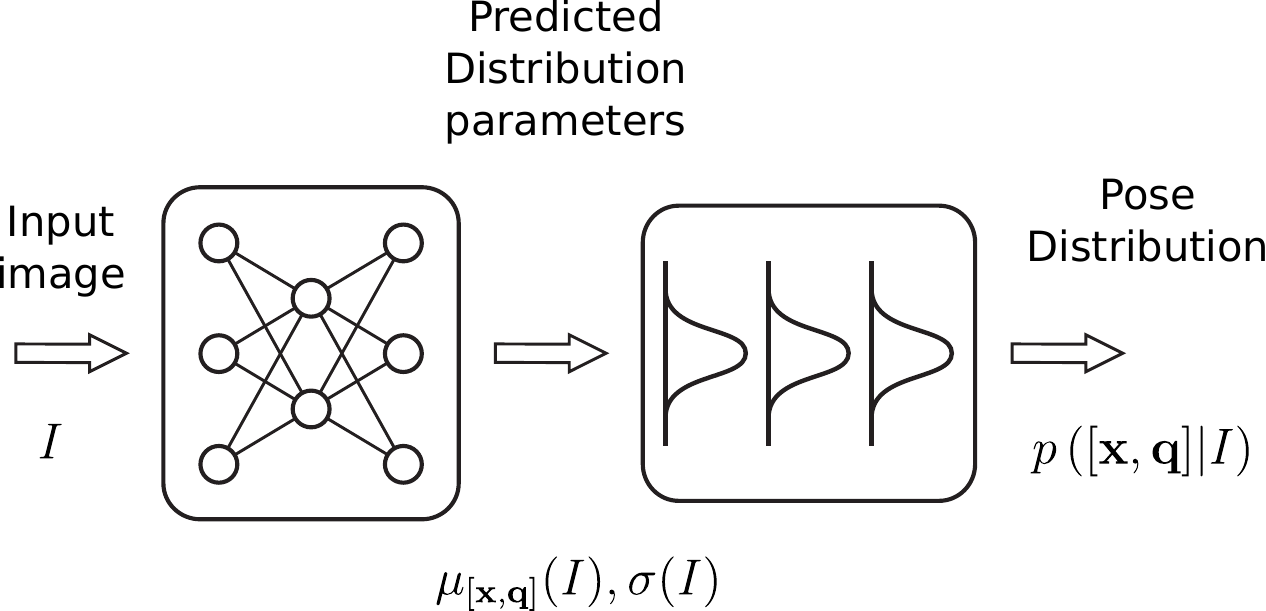}
    \caption{A mixture density network for modelling the multi-modal pose distribution \cite{bishop1994mixture}.}
    \label{fig:msn}
\end{figure}

\begin{equation}
p\left([\mathbf{x,q}]|I \right) = \sum_{i=1}^M \alpha_i(I)\mathcal{N}_i\left(\mu_{[\mathbf{x,q}]}(I),\sigma(I) \right)
\end{equation}
where $\mathcal{N}_i\left(\mu_{[\mathbf{x,q}]},\sigma |I\right)$ is a mixture component and $\alpha_i$ are the coefficients of the mixture distribution which satisfy the constraint $\sum_i \alpha_i = 1$. The mixing components are a function of the input image which is modelled by the network. As in the single Gaussian case, the network is trained to maximize the likelihood of the training data.



\section{Experiments}

In this section, the proposed approach is evaluated on outdoor and indoor datasets by comparing with the state-of-the-art methods.

\subsection{Datasets}

	Two well-known public datasets are employed in our experiments. They demonstrate indoor human motion and outdoor autonomous car driving scenarios, respectively. 
	
	Microsoft 7-Scenes Dataset which contains RGB-D image sequences of 7 different indoor environments \cite{shotton2013scene} was created by using a Kinect sensor. It has been widely used for camera tracking and relocalization \cite{kendall2015posenet}. The images were captured at $640\times480$ resolution with ground truth from KinectFusion system. Since there are several image sequences of one scene and each sequence is composed of about 500-1000 image frames, it is ideal for our experiments. Ground truth camera poses for the dataset are obtained using the KinectFusion algorithm \cite{newcombe2011kinectfusion} to produce smooth camera tracks and a dense 3D model of each scene. In our experiments, all the 7 scenes are adopted to evaluate the proposed method. We use the same Train and Test split of the sequences as used in the original paper. This dataset consists of both RGB and depth images. Although we focus mainly on RGB-only localization, our method extends naturally to the RGB-D case. 
	
	In order to further test the performance in large-scale outdoor environments, the recently released Oxford RobotCar dataset \cite{RobotCarDatasetIJRR} is used. It was recorded by using an autonomous Nissan LEAF car traversing in the central Oxford for a year period. The dataset contains high-resolution images from a Bumblebee stereo camera, LiDAR scanning, and GPS/INS. Since different weather conditions, such as sunny and snowy days, are exhibited in the dataset, it is very challenging for some tasks based on vision, e.g., global localization and loop closure detection across long terms and seasons. Because global re-localization does not need to have high-frequency images, the frame rate is about 1Hz in our robotcar experiments.
	
\subsection{Competing algorithms}
We compare our approach to the current state-of-the-art monocular camera localization methods. 

\noindent \textbf{Smoothing baseline}
The traditional means of integrating temporal information is to perform a filtering or smoothing operation on the independent pose predictions for each frame. We thus compare our method to a smoothing operation in order to investigate the advantage of using an RNN to capture the temporal information and whether global pose accuracies obtained for each frame are indeed more accurate than independent pose predictions. For our smoothing baseline, we use the spline fitting approach as per \cite{kroeger2014video}.

\noindent \textbf{Posenet}
Posenet uses a CNN to predict the pose of an input RGB image. The Posenet network is the GoogleNet architecture with the top-most fully connected layer removed and replaced by one with a 7-dimensional output and trained to predict the pose of the image. 

\noindent \textbf{Score-Forest}
The Score-Forest \cite{shotton2013scene} approach trains a random regression forest to predict the scene coordinates of pixels in the images. A set of predicted scene coordinates is then used to determine the camera pose using a RANSAC-loop. We use the open source implementation for our experiments\footnote{https://github.com/ISUE/relocforests}.

\subsection{Experiments on Microsoft 7-Scenes Dataset}

In this section we describe the experiments that we performed on the Microsoft 7-Scenes dataset.
	
\begin{table*}[h!]
\centering
\caption{Comparison to state-of-the-art approaches to monocular camera localization}
\label{tbl:accuracy}
\begin{tabular}{llll|lllllll}
\multicolumn{1}{c}{Scene} & \multicolumn{2}{c}{Frames} & \multicolumn{1}{c|}{\multirow{2}{*}{\begin{tabular}[c]{@{}c@{}}Spatial\\ Extent\end{tabular}}} & \multicolumn{1}{c}{\multirow{2}{*}{\begin{tabular}[c]{@{}c@{}}Score\\ Forest\end{tabular}}} & \multicolumn{1}{c}{\multirow{2}{*}{Posenet}} & \multicolumn{1}{c}{\multirow{2}{*}{\begin{tabular}[c]{@{}c@{}}Bayesian\\ Posenet\end{tabular}}} & \multicolumn{1}{c}{\multirow{2}{*}{\begin{tabular}[c]{@{}c@{}}Smoothing\\ Baseline\end{tabular}}} & \multicolumn{1}{c}{\multirow{2}{*}{VidLoc}} & \multicolumn{1}{c}{{VidLoc}} &\multicolumn{1}{c}{{VidLoc}}\\
 & Train & Test & \multicolumn{1}{c|}{} & \multicolumn{1}{c}{} & \multicolumn{1}{c}{} & \multicolumn{1}{c}{} & \multicolumn{1}{c}{} & \multicolumn{1}{c}{}& \multicolumn{1}{c}{RGB-D} & \multicolumn{1}{c}{Depth} \\ \hline
Chess       & 4000 & 2000 & 3x2x1m      & 0.03m & 0.32m & 0.37m & 0.32m     & 0.18m     & 0.16m     & 0.19m\\
Office      & 6000 & 4000 & 2.5x2x1.5m  & 0.04m & 0.48m & 0.48m & 0.38m     & 0.26m     & 0.24m     &0.32m\\
Fire        & 2000 & 2000 & 2.5x1x1m    & 0.05m & 0.47m & 0.43m & 0.45m     & 0.21m     &0.19m      &0.22m\\
Pumpkin     & 4000 & 2000 & 2.5x2x1m    & 0.04m & 0.47m & 0.61m & 0.42m     & 0.36m     & 0.33m     &0.15m    \\
Red kitchen & 7000 & 5000 & 4x3x1.5m    & 0.04m & 0.59m & 0.58m &  0.57m    & 0.31m     &0.28m      &0.38m\\
Stairs      & 2000 & 1000 & 2.5x2x1.5m  & 0.32m & 0.47m & 0.48m & 0.44m     &  0.26m    &0.24m      &0.27m\\
Heads       & 1000 & 1000 & 2x0.5x1m    & 0.06m & 0.29m & 0.31m & 0.19m     & 0.14m     &0.13m      &0.27m\\ \hline
Average     &  &  &  &  &  &  &  &  &\\ \hline
\end{tabular}
\end{table*}

The results of our experiments testing the accuracy of our method are shown in Table \ref{tbl:accuracy}. The proposed method significantly outperforms the Posenet approach in all of the test scenes, resulting in a $23.4\% - 55\%$ increase in accuracy. The SCoRe-forest outperforms the the RGB-only VidLoc. However, this is strictly not a fair comparison for two reasons: firstly, SCoRe-forest requires depth images as input; secondly, the SCoRe-forest sometimes produces pose estimates with gross errors although these are rejected by the RANSAC-loop, which means that pose estimates are not available for all frames. In contrast, our method produces reliable estimates for the entire sequence.

\begin{figure}[h!]
    \centering
    \includegraphics[width=\columnwidth]{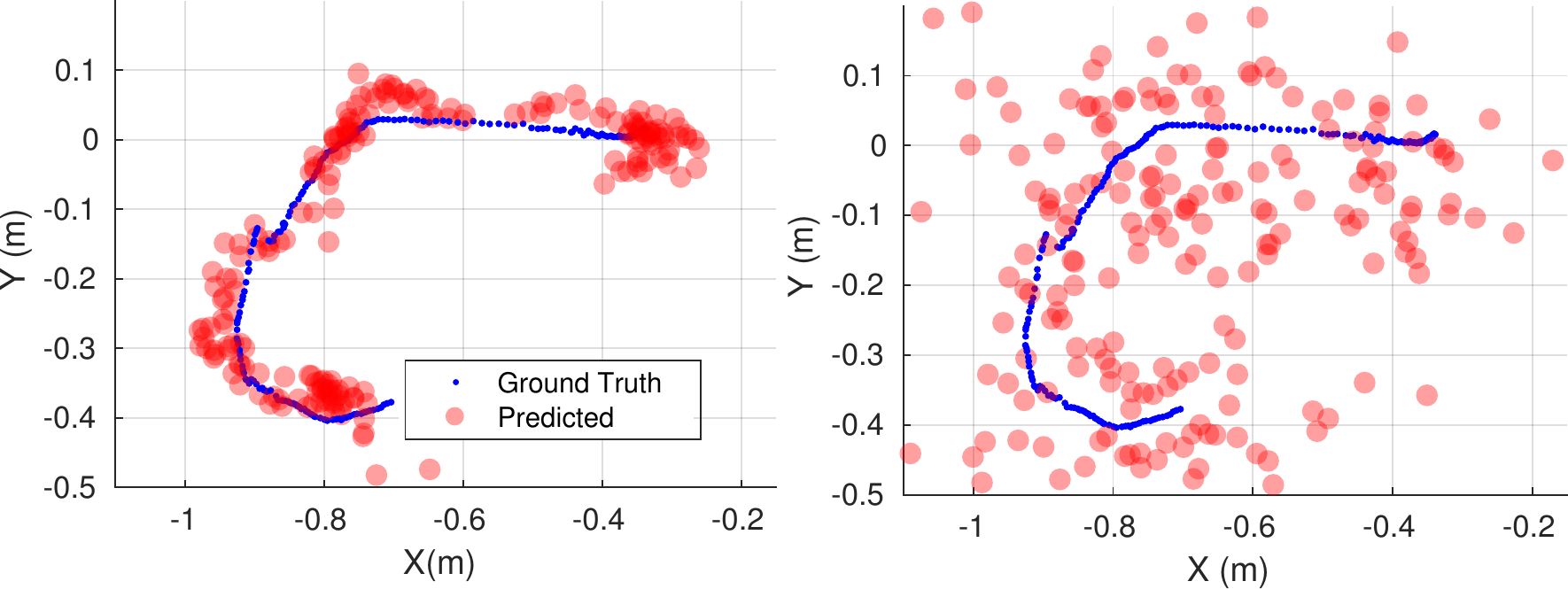}
    \caption{Comparison between our proposed method (left) and Posenet \cite{kendall2015posenet} (right). Our proposed method learns to estimate temporally-smooth 6-DoF poses from an input image stream.}
    \label{fig:my_label}
\end{figure}

\begin{figure}[h!]
	    \centering
	    \includegraphics[width=\columnwidth]{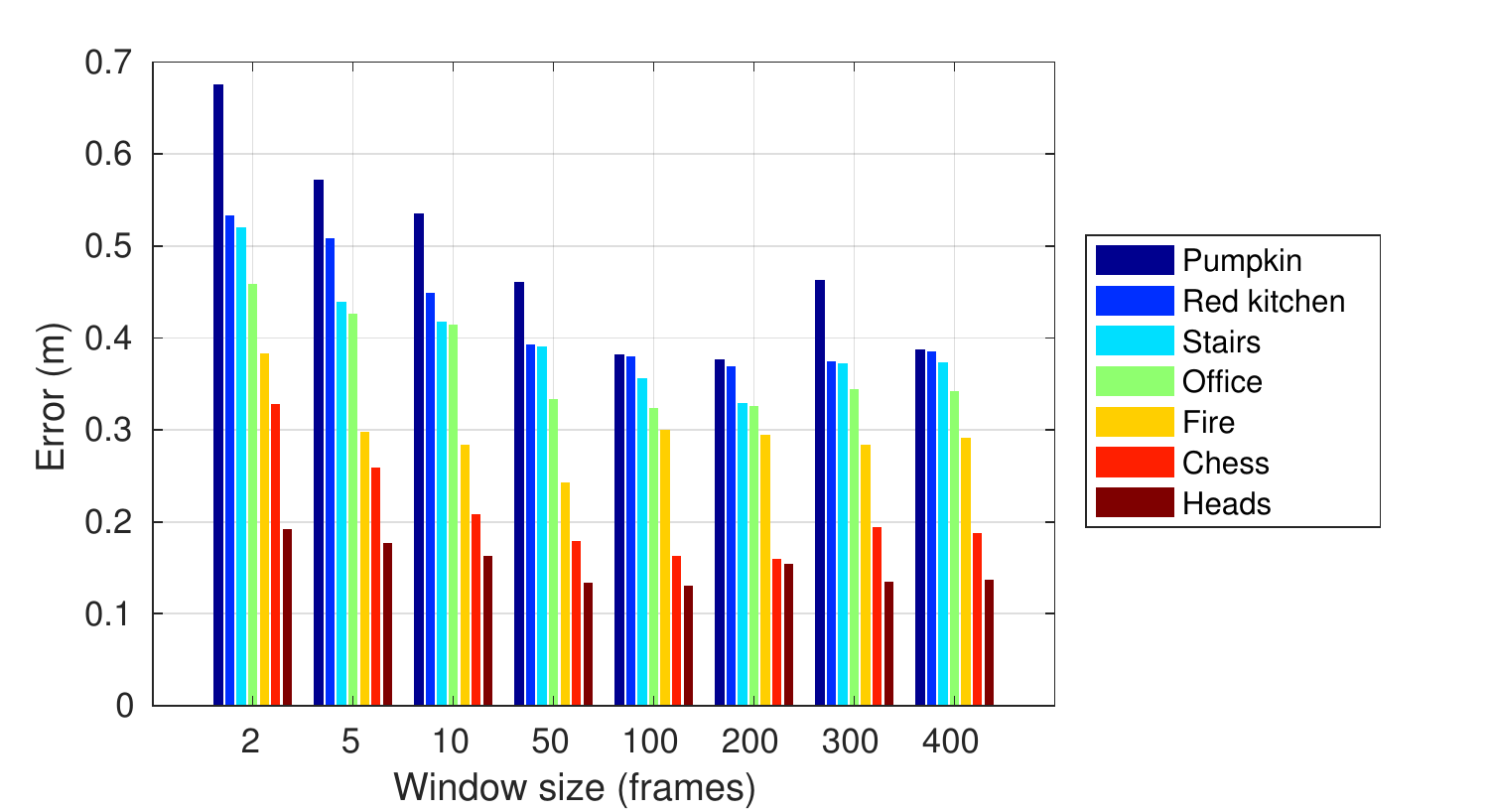}
	    \caption{The effect of window length on pose accuracy for the sequences in the Microsoft 7-Scenes dataset.}
	    \label{fig:ExpSeq}
	\end{figure}

\noindent We tested our method using both depth and RGB input and although our method seamlessly utilises the depth images when available, a disadvantage is that it cannot utilise the depth information to the extent that the SCoRe-Forest is able to. This is evidenced in the accuracy results reported in Table \ref{tbl:accuracy} where it can be seen that although our method consistently achieves centimeter accuracy, it does not outperform the SCoRe-Forest. This is surprising but perhaps indicative of the operation of the network. This suggests that the network learns to perform pose prediction in a similar fashion to an appearance based localization method. In this manner, it uses both the RGB and the depth information in the same way. This is in contrast to the SCoRe-forest approach where the depth information is explicitly used in a geometric pose computation by means of the PnP algorithm.
We note, however, that our method still has the advantage of being able to operate on RGB data when no depth information is available and is able to produce global pose estimates for all frames whereas the SCoRe-forest cannot.
\\
\noindent \textbf{Effect of sequence length}
A key result of this paper is shown in Figure \ref{fig:ExpSeq} which depicts the localization error as a function of the sequence length used. We have trained the models using sequence lengths of 200 frames in order to test the ability of the model to generalize to longer sequences. In all cases we ensure that the error is averaged over the same number and an even distribution across the test sequence. As expected, increasing the number of frames improves the localization accuracy. We also see that the model is able to generalize to longer sequence (i.e we still get an improvement in accuracy for sequence lengths greater than 200). At very long sequence lengths we experience diminishing returns - however this is not necessarily a product of the models inability to use this data but rather the actual utility of very long-term dependencies in predicting the current pose.

\noindent \textbf{Timings}
Our approach improves on the accuracy of Posenet, yet has very little impact on the computational time. This is because processing each frame only relies on the hidden state of the RNN from the previous time instance and image data of the current frame. Predicting a pose thus only requires a forward pass of the image through the CNN and propagating the hidden state. On our test machine with a Titan X Pascal GPU, this takes only 18ms using GoogleNet and 43ms using a VGG16 CNN. An interesting observation from our experiments is that the training time to create a usable localization network using the fine-tuning approach with Imagenet initialization is actually rather short. Typically convergence time (to around $90\%$) of final accuracy on the test data is around 50s.

\begin{figure}[h!]
	    \centering
	        \includegraphics[width=\columnwidth]{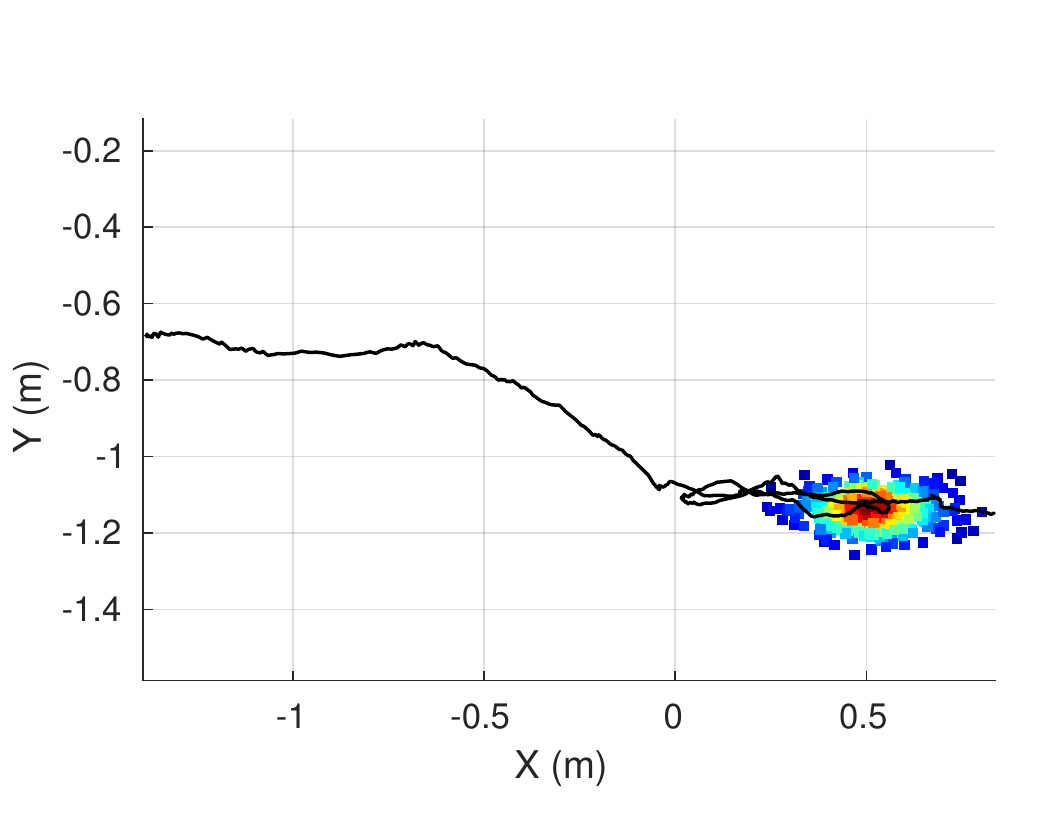}

	    \caption{Pose prediction density example on the ``Stairs'' sequence.}
	    \label{fig:ExpUncertainty}
	\end{figure}

\noindent \textbf{Uncertainty output} The 7-Scenes indoor dataset is extremely challenging, mainly due to the problem of perceptual aliasing as shown in Figure \ref{fig:Exp7ScenesImages}. One image was taken from the bottom of the staircase while the other was taken near the top. The multi-modal global pose prediction of a section of the ``Stairs'' sequence is shown in Figure \ref{fig:ExpUncertainty} which was produced using a mixture model of 3 Gaussians. From the figure it is evident that the predicted distribution adaquately. However, in many cases we found that the predicted variance is rather high and we leave it as future work to improve the variance prediction.

\subsection{Experiments on RobotCar Dataset}

	The experiments on the Oxford RobotCar Dataset are given in this section. Since the GPS/INS poses are relatively noisy (zig-zag track), they are fused with stereo visual odometry by using pose graph SLAM to produce smooth ground truth for training. In our experiments, three image sequences are used for training, while the trained models are tested on another new testing sequence. 
	
	\begin{figure}[h!]
	    \centering
	    \begin{subfigure}[c]{\columnwidth}
	        \includegraphics[width=\textwidth]{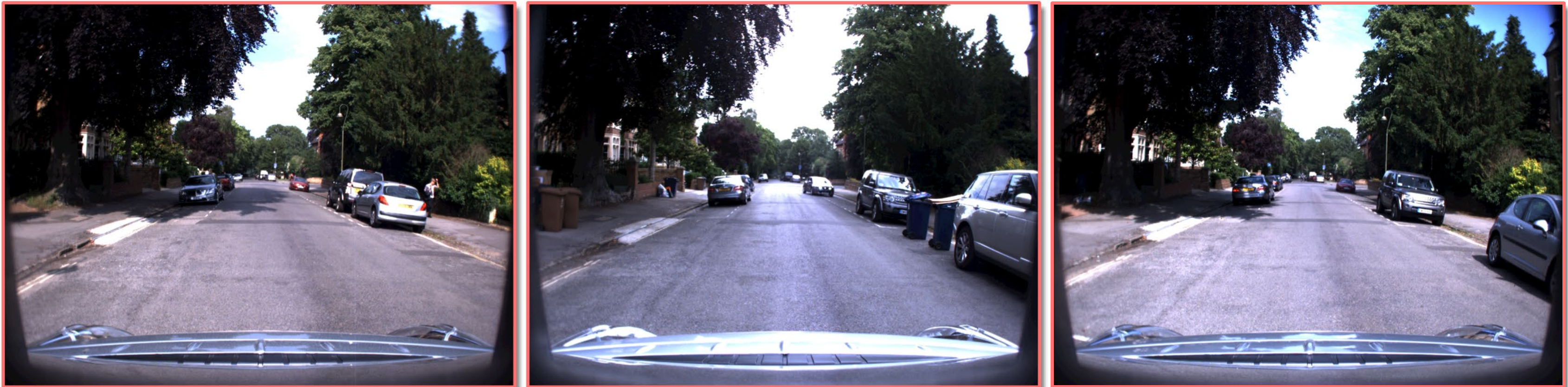}
	        \caption{Images on a same location at different times.}
	        \label{fig:ExpCarSamePoint}
	    \end{subfigure}\\
	    \begin{subfigure}[c]{\columnwidth}
	        \includegraphics[width=\textwidth]{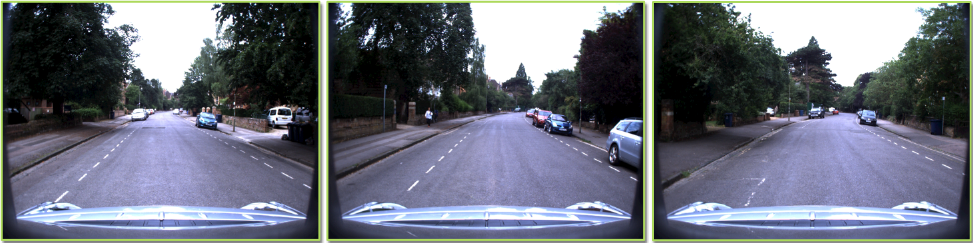}
	        \caption{Images on different locations but close times.}
	        \label{fig:ExpCarDiffPoints}
	    \end{subfigure}
	    \caption{Images of the RobotCar dataset to show limited appearance distinction with dynamic changes on a same location and perceptual aliasing among different locations.}
	    \label{fig:ExpCarImages}
	\end{figure}

	\begin{figure*}[t!]
		\centering
		\begin{subfigure}[c]{0.39\textwidth}
			\includegraphics[width=\textwidth]{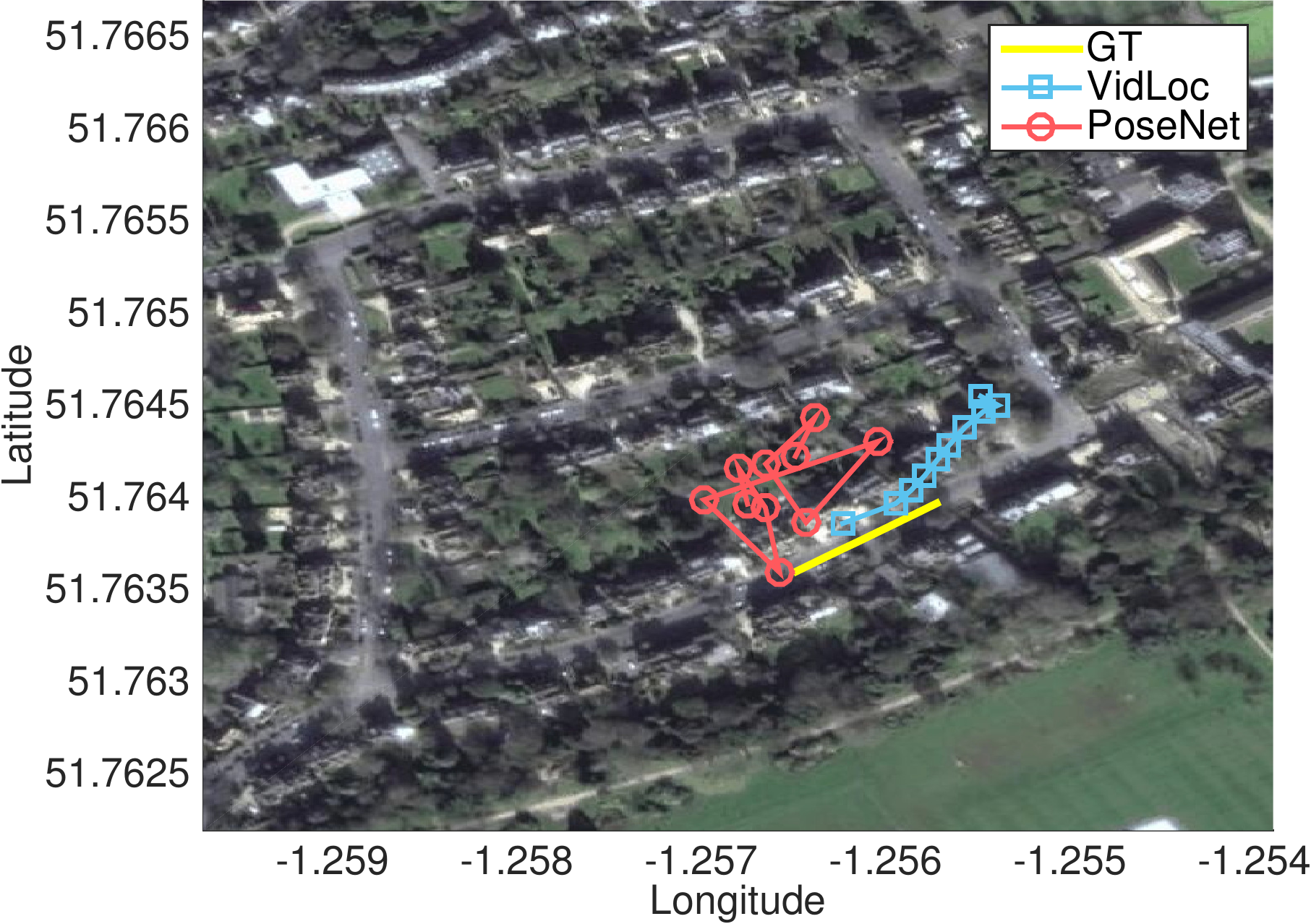}
			\caption{Sequence length: 10.}
			\label{fig:ExpCarTrajsMapSeq10}
		\end{subfigure}
		\begin{subfigure}[c]{0.39\textwidth}
			\includegraphics[width=\textwidth]{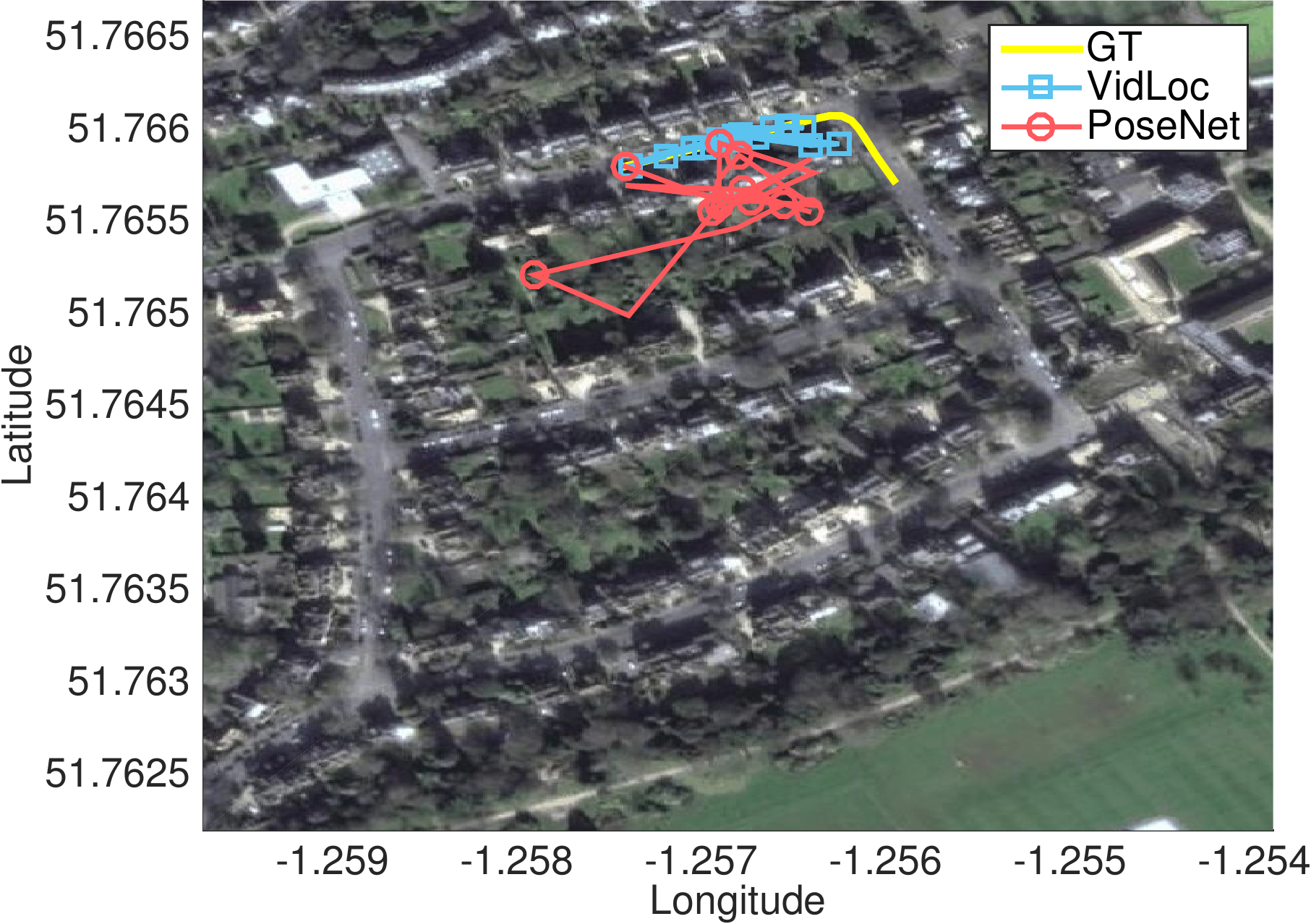}
			\caption{Sequence length: 20.}
			\label{fig:ExpCarTrajsMapSeq20}
		\end{subfigure}\\
		\begin{subfigure}[c]{0.39\textwidth}
			\includegraphics[width=\textwidth]{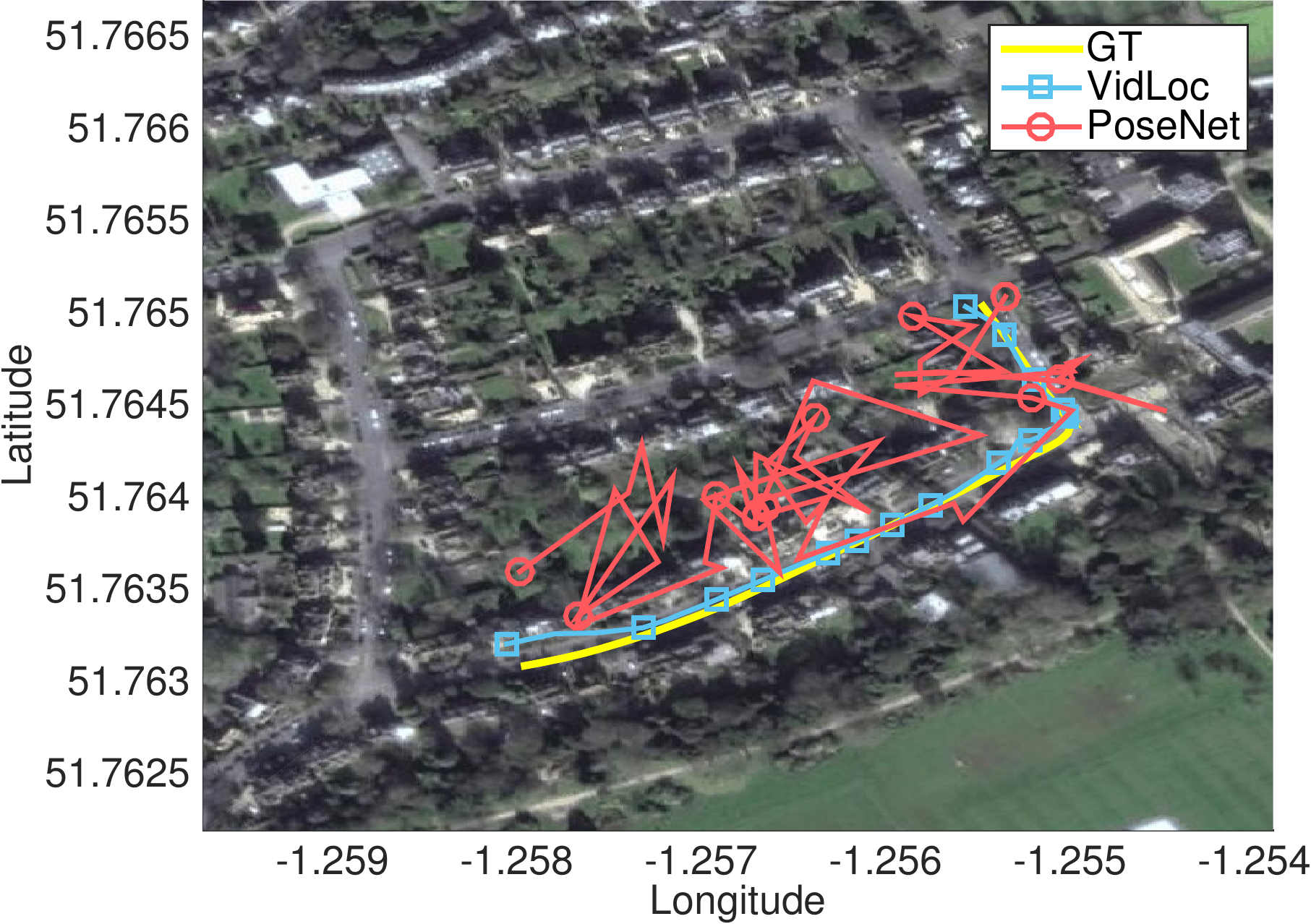}
			\caption{Sequence length: 50.}
			\label{fig:ExpCarTrajsMapSeq50}
		\end{subfigure}
		\begin{subfigure}[c]{0.39\textwidth}
			\includegraphics[width=\textwidth]{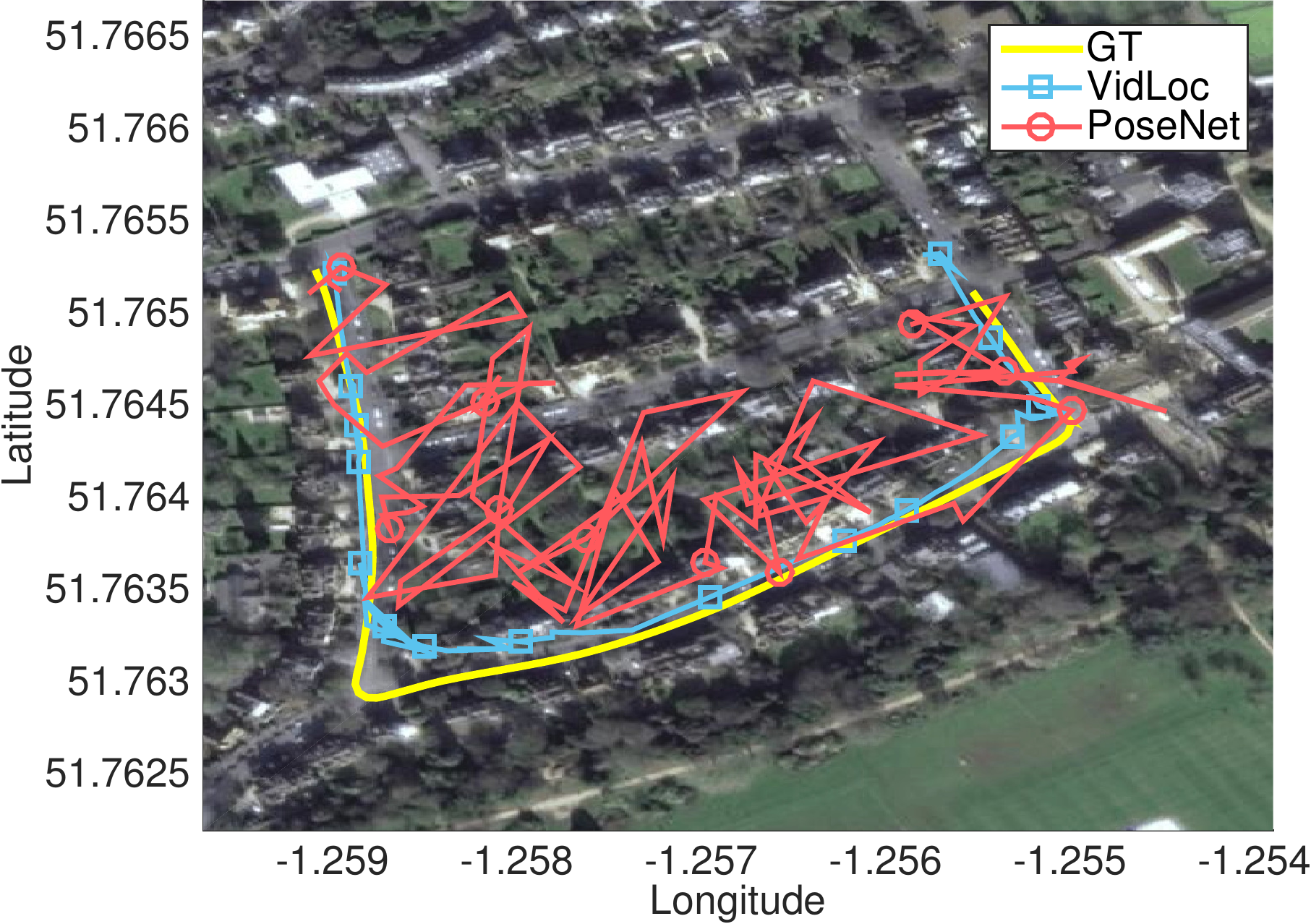}
			\caption{Sequence length: 100.}
			\label{fig:ExpCarTrajsMapSeq100}
		\end{subfigure}
		\caption{Global localization results on different lengths of sequences superimposed on Google Map.}
		\label{fig:ExpCarTrajsMap}
	\end{figure*}
	
	The image sequences selected are very challenging for global re-localization. As shown in \figurename{\ref{fig:ExpCarImages}}, the images are mostly filled with roads and trees, which do not have distinct and consistent appearance features. Specifically, three images of a same location yet captured at different times are presented in \figurename{\ref{fig:ExpCarSamePoint}}. Although they are taken at a same position, the cars parking along the road introduce significant appearance changes. Without viewing the buildings around, the only consistent objects which can be useful for global re-localization are the trees and roads. However, they are subtle in terms of image context. For example, \figurename{\ref{fig:ExpCarDiffPoints}} shows sample images of three different locations which share very similar appearance. This perceptual aliasing makes global re-localization more challenging by only using one single image.

	The global re-localization results of the testing image sequence with lengths 10, 20, 50 and 100 are shown in \figurename{\ref{fig:ExpCarTrajsMap}} against ground truth. They are also superimposed on Google Map. It can be seen that the result of the proposed method improves as the length of the sequence increases, and the re-localization results of the lengths 50 and 100 match with the roads consistently. It is interesting to see that its trajectories are also able to track the shape of motion by end-to-end learning. In contrast, the Posenet which uses a single image suffers from noisy pose estimates around the ground truth. This experiment validates the effectiveness and necessary of using sequential images for global re-localization, mitigating the problems of perceptual aliasing and improving localization accuracy.

\begin{figure}
		\centering
		\begin{subfigure}[c]{0.5\columnwidth}
			\includegraphics[width=\textwidth]{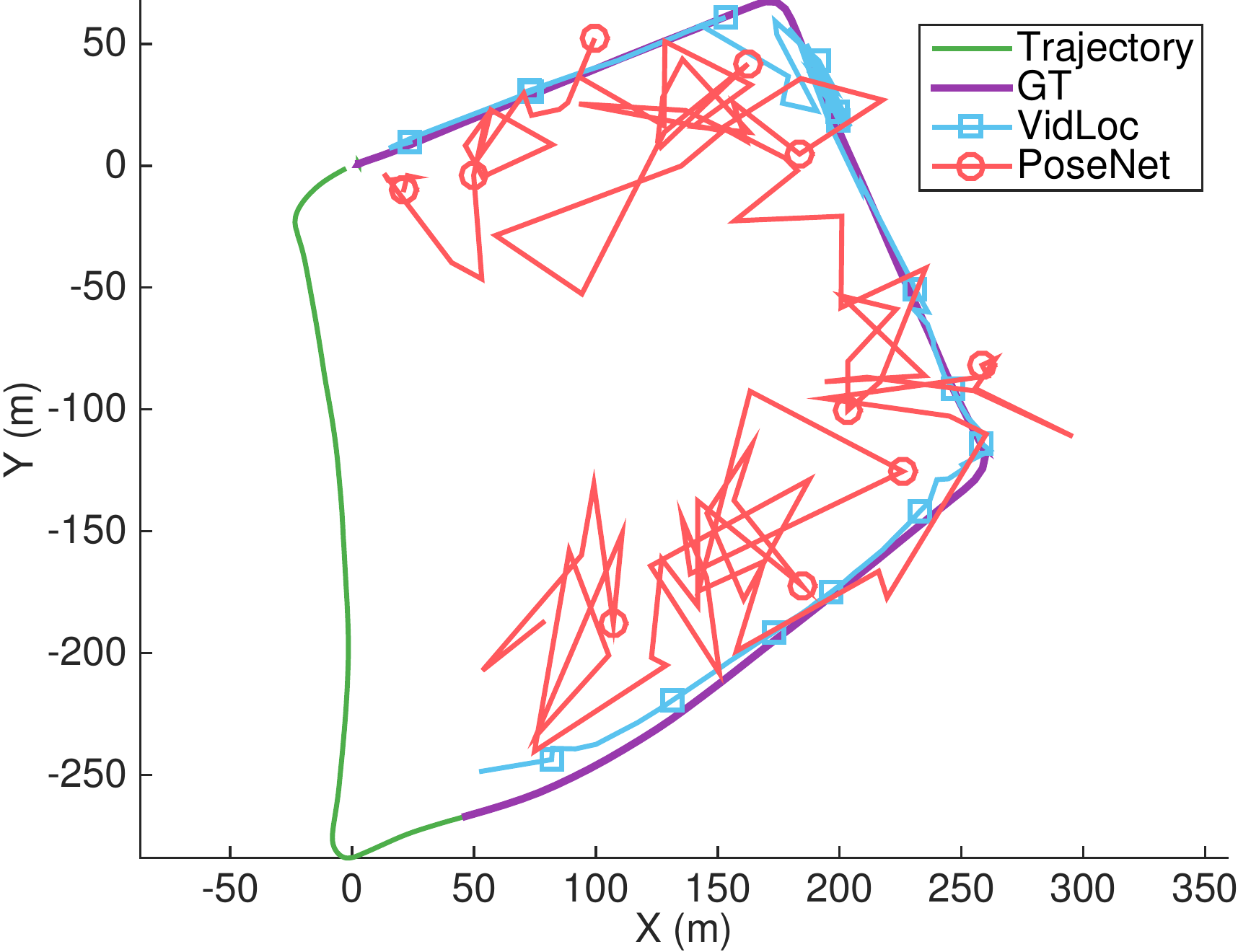}
			\caption{Localization results.}
			\label{fig:ExpCarTraj}
		\end{subfigure}
		\begin{subfigure}[c]{0.49\columnwidth}
			\includegraphics[width=\textwidth]{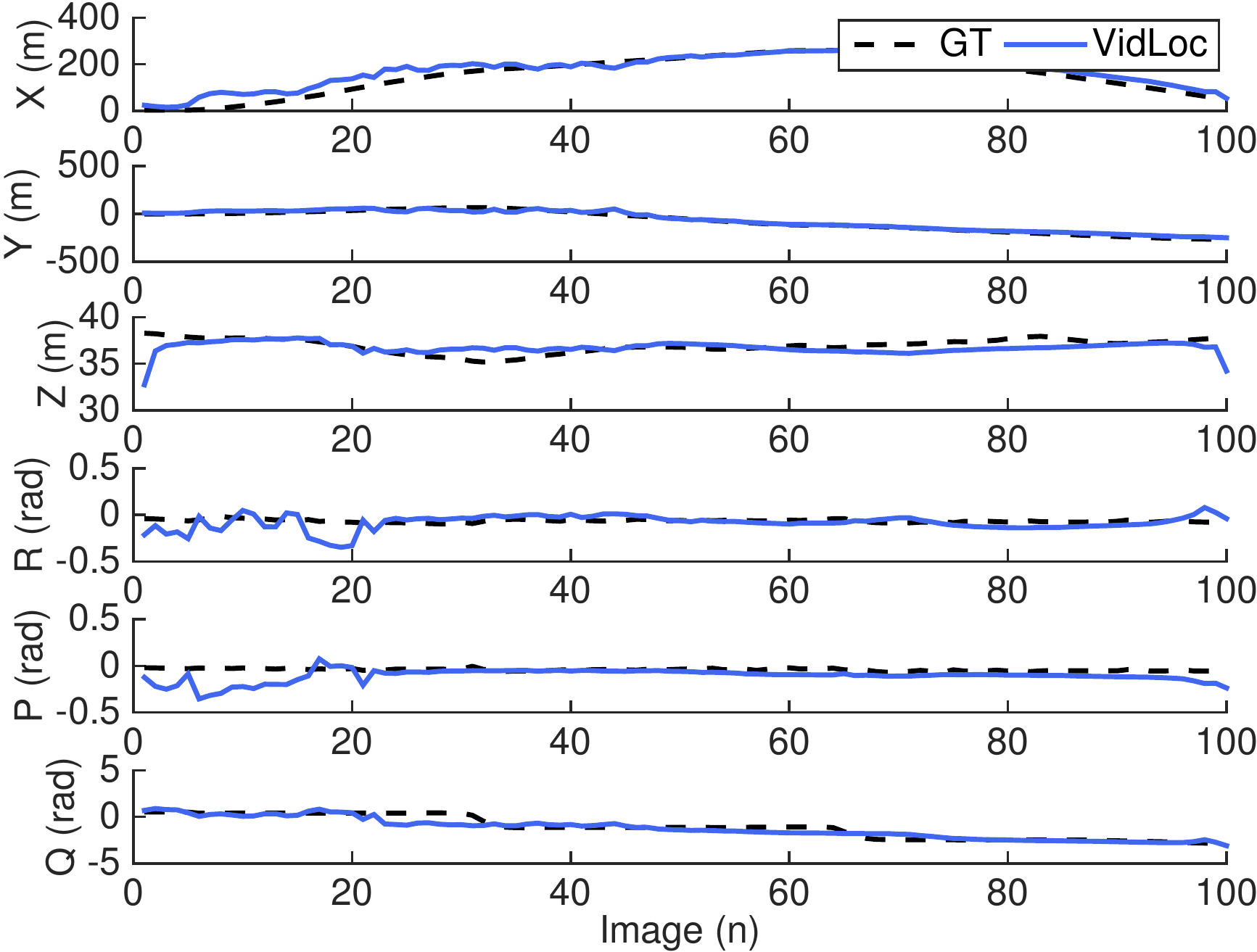}
			\caption{Estimates of 6-DoF poses.}
			\label{fig:ExpCarTraj_Pose}
		\end{subfigure}
		\caption{Localization result and estimates of 6-DoF poses of a sequence with 100 length.}
		\label{fig:ExpCarExample}
	\end{figure}

	Localization trajectories and 6-DoF pose estimation of a sequence with 100 length are given in \figurename{\ref{fig:ExpCarExample}}. It further shows that the localization result is smooth and accurate. The corresponding estimation of the 6-DoF poses on x, y, z, roll, pitch and yaw is described in \figurename{\ref{fig:ExpCarTraj_Pose}}. It can be seen that the proposed method can track the ground truth accurately in terms of 6-DoF pose estimation. This is of importance when using the localization result for re-localization and loop closure detection. 
	
\begin{figure}
		\centering
		\begin{subfigure}[c]{0.5\columnwidth}
			\includegraphics[width=\textwidth]{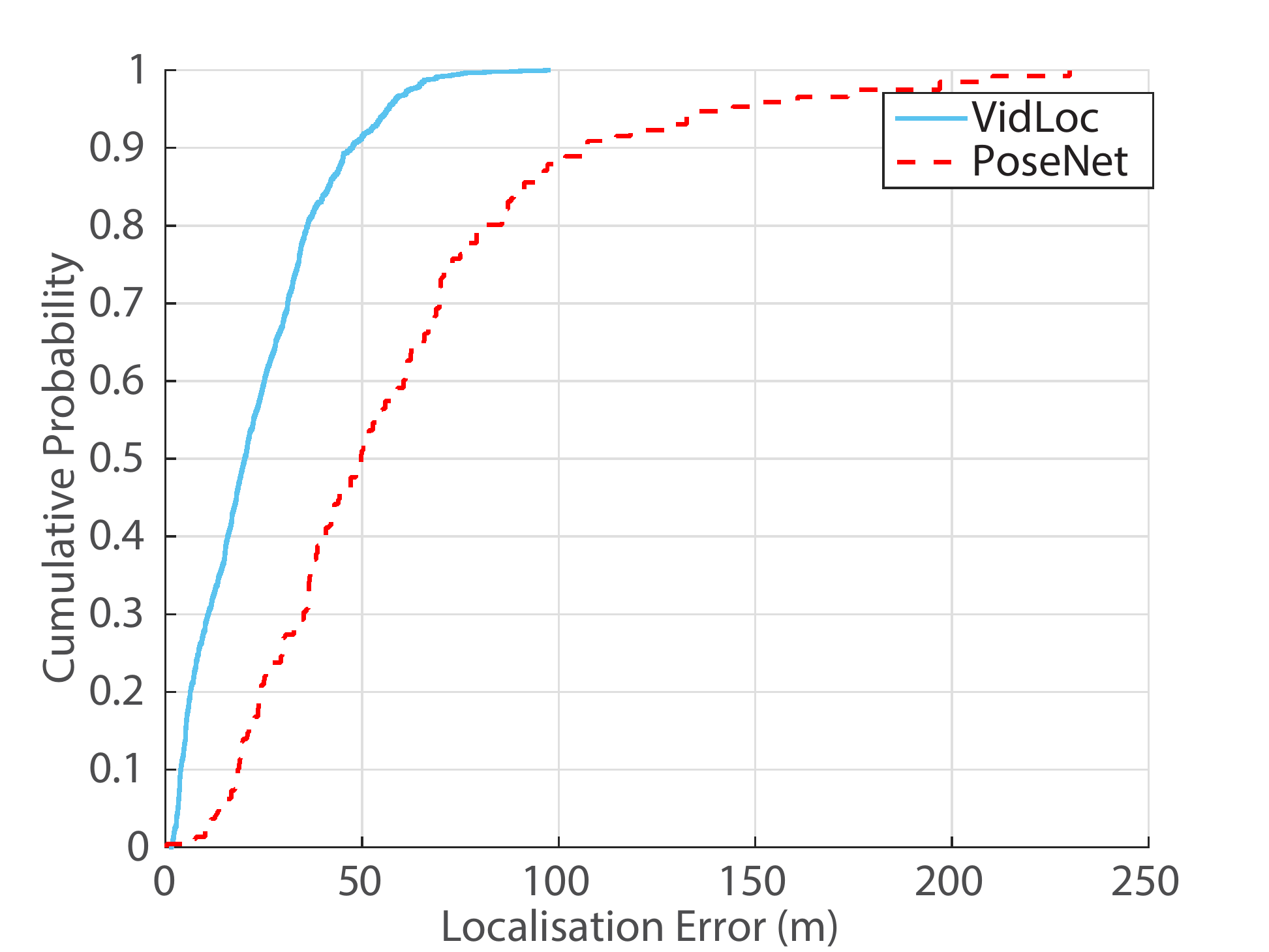}
			\caption{Distribution of errors.}
			\label{fig:ExpCarCDFSeq100}
		\end{subfigure}
		\begin{subfigure}[c]{0.49\columnwidth}
			\includegraphics[width=\textwidth]{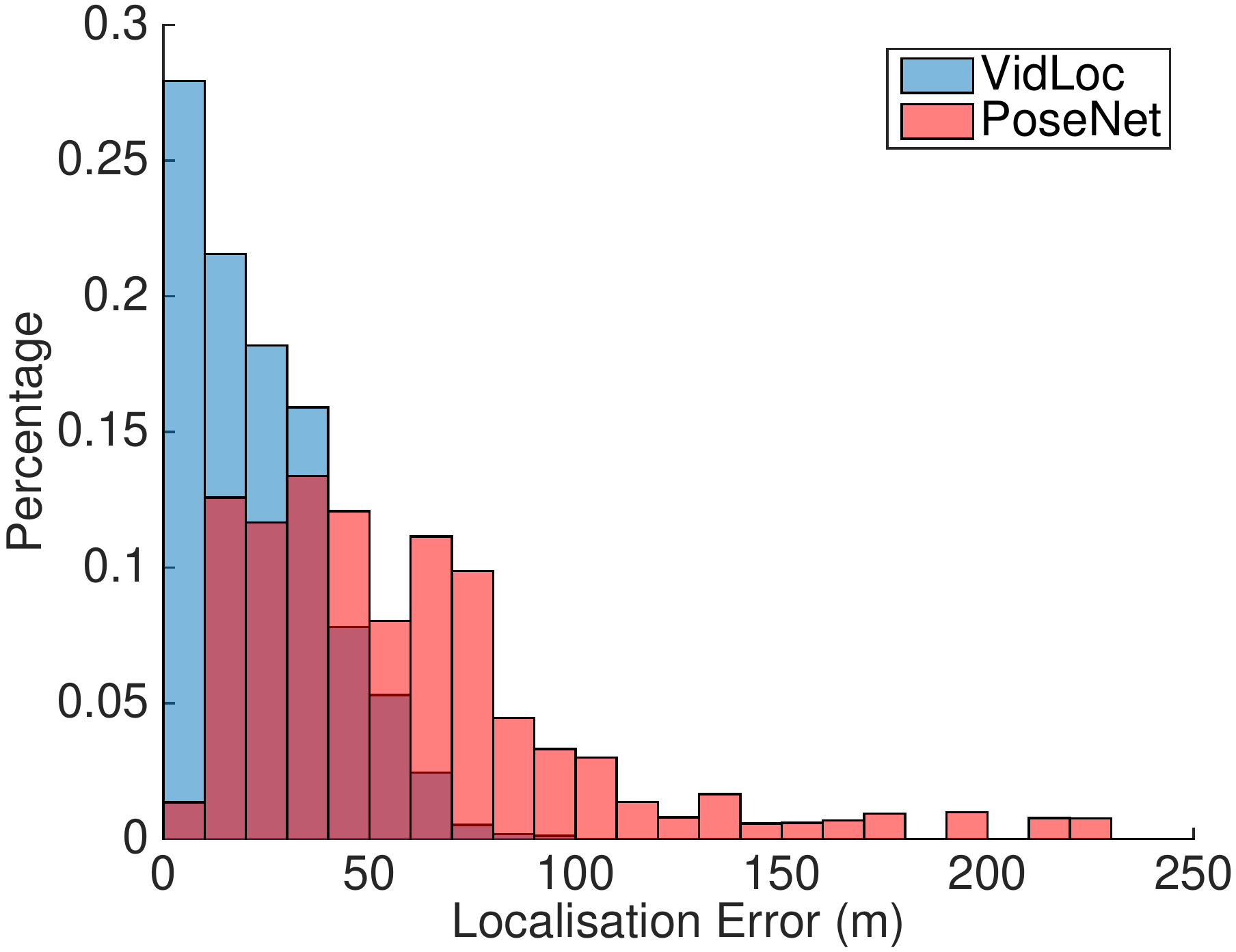}
			\caption{Histogram of errors.}
			\label{fig:ExpCarHistSeq100}
		\end{subfigure}
		\caption{Distribution and histogram of localization errors of all sequences with 100 length.}
		\label{fig:ExpCarSeq100}
	\end{figure}

	\figurename{\ref{fig:ExpCarSeq100}} illustrates the distribution and histogram of the re-localization errors (mean squared errors) of all sequences with 100 length. Statistically more than half of poses estimated by the proposed method are within $20$ meters, while this is less than $15\%$ percentage for Posenet. Moreover, there are some big errors, e.g., more than 200 meters, of Posenet, which indicates that it may have perceptual aliasing problems during pose estimation. It tends to be common in this challenging dataset, as shown in \figurename{\ref{fig:ExpCarImages}}. Therefore, it is verified that the recurrent model encapsulating the relationship between consecutive image frames is effective for global re-localisation using a video clip.

\section{Conclusion}

We have presented an approach for 6-DoF video-clip re-localization that exploits the temporal smoothness of the video stream to improve the localization accuracy of the global pose estimates. We have studied the impact of window size and shown that our method outperforms the closest related approaches for monocular RGB localization by a fair margin. 
	
For future work we intend to investigate means of making better use of the depth information, perhaps by forcing the network to learn to make use of geometrical information. One means of doing this would be to try predict the scene coordinates of the input RGB-D image using the CNN in an intermediate layer and then derive the pose from this and the input image. In essence this would be like unifying appearance-based localization and geometry-based localization in one model.

\balance

{\small
\bibliographystyle{ieee}
\bibliography{egbib}

\begin{thebibliography}{10}\itemsep=-1pt

\bibitem{bianco2015logo}
S.~Bianco, M.~Buzzelli, D.~Mazzini, and R.~Schettini.
\newblock Logo recognition using cnn features.
\newblock In {\em International Conference on Image Analysis and Processing},
  pages 438--448. Springer, 2015.

\bibitem{bishop1994mixture}
C.~M. Bishop.
\newblock Mixture density networks.
\newblock 1994.

\bibitem{clark17vinet}
R.~Clark, S.~Wang, H.~Wen, A.~Markham, and N.~Trigoni.
\newblock Vinet: Visual-inertial odometry as a sequence-to-sequence learning
  problem.
\newblock In {\em Proceedings of the Thirty-First {AAAI} Conference on
  Artificial Intelligence}, 2017.

\bibitem{gal2015bayesian}
Y.~Gal and Z.~Ghahramani.
\newblock Bayesian convolutional neural networks with bernoulli approximate
  variational inference.
\newblock {\em arXiv preprint arXiv:1506.02158}, 2015.

\bibitem{goldstein2016shapefit}
T.~Goldstein, P.~Hand, C.~Lee, V.~Voroninski, and S.~Soatto.
\newblock Shapefit and shapekick for robust, scalable structure from motion.
\newblock In {\em European Conference on Computer Vision}, pages 289--304.
  Springer, 2016.

\bibitem{gustafsson2002particle}
F.~Gustafsson, F.~Gunnarsson, N.~Bergman, U.~Forssell, J.~Jansson, R.~Karlsson,
  and P.-J. Nordlund.
\newblock Particle filters for positioning, navigation, and tracking.
\newblock {\em IEEE Transactions on signal processing}, 50(2):425--437, 2002.

\bibitem{hochreiter1997long}
S.~Hochreiter and J.~Schmidhuber.
\newblock Long short-term memory.
\newblock {\em Neural computation}, 9(8):1735--1780, 1997.

\bibitem{kendall2015modelling}
A.~Kendall and R.~Cipolla.
\newblock Modelling uncertainty in deep learning for camera relocalization.
\newblock {\em Proceedings of the International Conference on Robotics and
  Automation ({ICRA})}, 2016.

\bibitem{kendall2015posenet}
A.~Kendall, M.~Grimes, and R.~Cipolla.
\newblock Posenet: A convolutional network for real-time 6-dof camera
  relocalization.
\newblock In {\em Proceedings of the IEEE International Conference on Computer
  Vision}, pages 2938--2946, 2015.

\bibitem{kroeger2014video}
T.~Kroeger and L.~Van~Gool.
\newblock Video registration to sfm models.
\newblock In {\em European Conference on Computer Vision}, pages 1--16.
  Springer, 2014.

\bibitem{RobotCarDatasetIJRR}
W.~Maddern, G.~Pascoe, C.~Linegar, and P.~Newman.
\newblock {1 Year, 1000km: The Oxford RobotCar Dataset}.
\newblock {\em The International Journal of Robotics Research (IJRR)}, to
  appear.

\bibitem{middelberg2014scalable}
S.~Middelberg, T.~Sattler, O.~Untzelmann, and L.~Kobbelt.
\newblock Scalable 6-dof localization on mobile devices.
\newblock In {\em European Conference on Computer Vision}, pages 268--283.
  Springer, 2014.

\bibitem{newcombe2011kinectfusion}
R.~A. Newcombe, S.~Izadi, O.~Hilliges, D.~Molyneaux, D.~Kim, A.~J. Davison,
  P.~Kohi, J.~Shotton, S.~Hodges, and A.~Fitzgibbon.
\newblock Kinectfusion: Real-time dense surface mapping and tracking.
\newblock In {\em Mixed and augmented reality (ISMAR), 2011 10th IEEE
  international symposium on}, pages 127--136. IEEE, 2011.

\bibitem{newson2009hidden}
P.~Newson and J.~Krumm.
\newblock Hidden markov map matching through noise and sparseness.
\newblock In {\em Proceedings of the 17th ACM SIGSPATIAL international
  conference on advances in geographic information systems}, pages 336--343.
  ACM, 2009.

\bibitem{sattler2011fast}
T.~Sattler, B.~Leibe, and L.~Kobbelt.
\newblock Fast image-based localization using direct 2d-to-3d matching.
\newblock In {\em 2011 International Conference on Computer Vision}, pages
  667--674. IEEE, 2011.

\bibitem{sattler2012improving}
T.~Sattler, B.~Leibe, and L.~Kobbelt.
\newblock Improving image-based localization by active correspondence search.
\newblock In {\em European Conference on Computer Vision}, pages 752--765.
  Springer, 2012.

\bibitem{schuster1997bidirectional}
M.~Schuster and K.~K. Paliwal.
\newblock Bidirectional recurrent neural networks.
\newblock {\em IEEE Transactions on Signal Processing}, 45(11):2673--2681,
  1997.

\bibitem{sharif2014cnn}
A.~Sharif~Razavian, H.~Azizpour, J.~Sullivan, and S.~Carlsson.
\newblock Cnn features off-the-shelf: an astounding baseline for recognition.
\newblock In {\em Proceedings of the IEEE Conference on Computer Vision and
  Pattern Recognition Workshops}, pages 806--813, 2014.

\bibitem{shotton2013scene}
J.~Shotton, B.~Glocker, C.~Zach, S.~Izadi, A.~Criminisi, and A.~Fitzgibbon.
\newblock Scene coordinate regression forests for camera relocalization in
  rgb-d images.
\newblock In {\em Proceedings of the IEEE Conference on Computer Vision and
  Pattern Recognition}, pages 2930--2937, 2013.

\bibitem{simonyan2014very}
K.~Simonyan and A.~Zisserman.
\newblock Very deep convolutional networks for large-scale image recognition.
\newblock {\em arXiv preprint arXiv:1409.1556}, 2014.

\bibitem{sunderhauf2015place}
N.~Sunderhauf, S.~Shirazi, A.~Jacobson, F.~Dayoub, E.~Pepperell, B.~Upcroft,
  and M.~Milford.
\newblock Place recognition with convnet landmarks: Viewpoint-robust,
  condition-robust, training-free.
\newblock {\em Proceedings of Robotics: Science and Systems XII}, 2015.

\bibitem{szegedy2015going}
C.~Szegedy, W.~Liu, Y.~Jia, P.~Sermanet, S.~Reed, D.~Anguelov, D.~Erhan,
  V.~Vanhoucke, and A.~Rabinovich.
\newblock Going deeper with convolutions.
\newblock In {\em Proceedings of the IEEE Conference on Computer Vision and
  Pattern Recognition}, pages 1--9, 2015.

\bibitem{zhang2006image}
W.~Zhang and J.~Kosecka.
\newblock Image based localization in urban environments.
\newblock In {\em 3D Data Processing, Visualization, and Transmission, Third
  International Symposium on}, pages 33--40. IEEE, 2006.

\bibitem{zhou2014learning}
B.~Zhou, A.~Lapedriza, J.~Xiao, A.~Torralba, and A.~Oliva.
\newblock Learning deep features for scene recognition using places database.
\newblock In {\em Advances in neural information processing systems}, pages
  487--495, 2014.

\end{thebibliography}
}

\end{document}